\documentclass[letterpaper]{article}
\usepackage{aaai}
\usepackage{times}
\usepackage{helvet}
\usepackage{courier}
\usepackage{graphicx}
\usepackage{times}
\usepackage{latexsym}
\usepackage{booktabs}
\usepackage{amsmath}
\usepackage{amssymb}
\usepackage{graphicx}
\usepackage{bm}
\usepackage{CJKutf8}
\usepackage{amsmath}
\usepackage{url}
\usepackage{tabularx}
\usepackage{subfig}
\usepackage{multirow}

\usepackage{color}
\usepackage[ruled,linesnumbered]{algorithm2e}
\begin{document}
%
\title{MetaOnce: A Metaverse Framework Based on Multi-scene Relations and Entity-relation-event Game}
\author{
Hongyin Zhu\\
Department of Computer Science and Technology, Tsinghua University, Beijing, China\\
zhuhongyin2020@mail.tsinghua.edu.cn\\
}
\maketitle
\begin{CJK*}{UTF8}{gbsn}
\begin{abstract}
Existing metaverse systems lack rich relation types between entities and events. The challenge is that there is no portable framework to introduce rich concepts, relations, events into the metaverse. This paper introduces a new metaverse framework, MetaOnce. This framework proposes to build multi-scene graphs. This framework not only describes rich relations in a single scene but also combines multiple scene graphs into a complete graph for more comprehensive analysis and inference. Prior social network systems mainly describe friend relations. They ignore the effect of entity-relation-event games on the metaverse system and existing rule constraints. We propose a rule controller and impose constraints on the relations that allow the framework to behave in a compliant manner. We build a metaverse system to test the features of the framework, and experimental results show that our framework can build a multi-scene metaverse with memory and rule constraints. 
\end{abstract}

\section{Introduction}
The metaverse system aims to map entities in the real world into the virtual world, thereby generating parallel virtual avatars. The blueprint for the metaverse \cite{sparkes2021metaverse} (Metaverse = Meta + Universe) is to build a digital twin of the real world. The concept of the metaverse is constantly expanding, such as game worlds, social networks, virtual reality (VR) or augmented reality (AR) technologies are all related to the metaverse. Building metaverses inevitably involves establishing relations and events. However, most systems usually describe friend relations or follower relations. They ignore three issues: (1) There are many types of relations between entities, for example, relations between people can be friends, spouses, lovers, enemies, competition, etc. (2) Different relations and events involved in different scenes (context). For example, in a classroom scene, the relations between people are classmates, teacher-student, etc., but in a game scene, the same people become teammates and opponents. (3) The game \cite{owen2013game,thomas2012games} between entities, relations, and events gives entities different powers. The capacity of an entity to control relations is different at different times, for example, when a person has a spouse, he/she cannot choose to marry another person.

\begin{figure}[htbp!]
\centering
\includegraphics[width=2.5in]{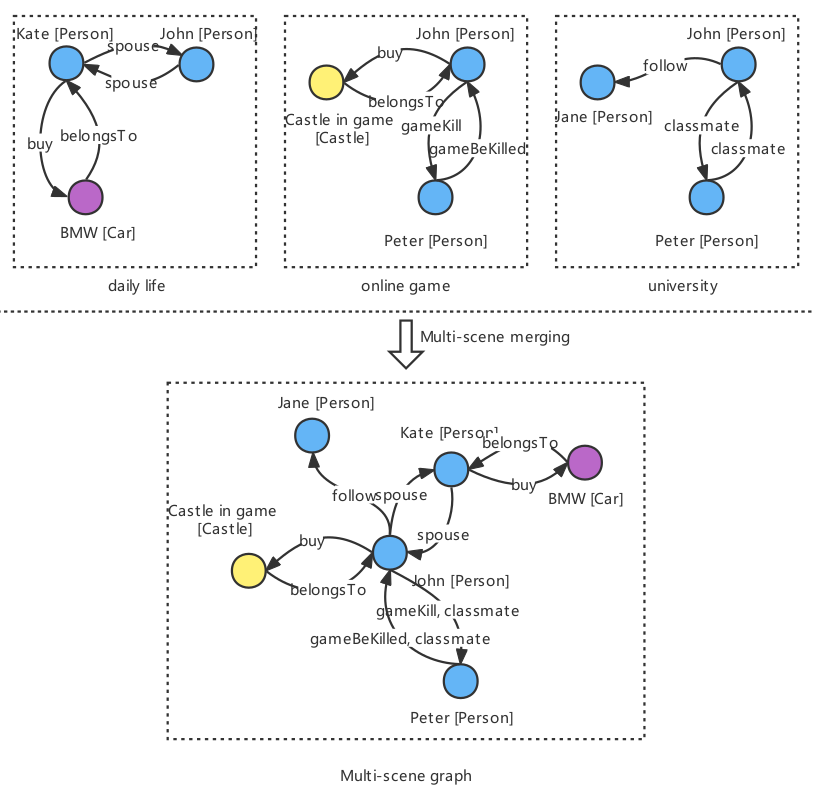}
\caption{An example of merging the relations of multiple scenes into a complete graph\label{example1}}
\end{figure}

Figure \ref{example1} uses an example to demonstrate the process of merging graphs created in 3 separate scenes (daily life, online game, university) into one complete graph.
Relations between two people are easy to change. For example, it is easy to change from lovers to spouses, but it is difficult to get back from spouse to the original relation. A married person cannot be a spouse with another person before divorce, and it is almost impossible for a divorced couple to remarry with each other within a certain time. No one else may buy a house belonging to this person until the person agrees to sell. Therefore, after the relation between entities is established, many constraints or rights will be imposed on the entity itself, and the capacity of the entity to control the relations at different times is also different. We hope to address the aforementioned issues. We propose the metaverse framework, \textbf{MetaOnce} (MetaOnce = Metaverse + Once). The name means that \textbf{once} a relation in the \textbf{metaverse} is established, changed, or cancelled, the event exists in the metaverse's memory and continues to affect the metaverse in the future. This also means that users are also responsible for their actions in the metaverse. Our metaverse framework can be used to create private, group, or Internet metaverses of varying scales. 

Maintaining the rich relations, multi-scene graph, and the entity-relation-event game is a non-trivial task. This task faces 3 challenges. The first challenge is to introduce the ontology for the metaverse, including predefining the rich types of concepts, relations, and events that the metaverse needs to contain. We propose to use open ontologies described in subsection Ontology of MetaOnce. As an illustrative example, we use the schema.org ontology as a demonstration.

The second challenge is the multi-scene graph problem. Only a certain range of relations and events are included in a particular scene. When moving to another scene, the relations and event types appropriate for the new scene also change. Besides, relations in different scenes cannot be jointly analyzed and inferred. To address this problem, we propose multi-scene graph merging. 

The third challenge is the entity-relation-event game. The capacity of an entity to control relation is not constant. In reality, once the relation between people changes, it is difficult to restore the original relations. People always tend to keep their relations up to date. In metaverse, we want to keep up-to-date relations between entities in real-time. We design some entity-relation-event rules for the framework to give the framework event memory. When an event occurs, it imposes constraints on the running rules and affects the metaverse. We propose different types of rule constraints for each relation, such as exclusive constraints, symmetry constraints, asymmetric co-occurrence constraints, mutual termination constraints, irreversible constraints, etc.

To evaluate the features of the framework, we construct a metaverse system and invite users to build metaverse examples and test building scene graphs, merging multi-scene graphs, and entity-relation-event rule controllers respectively. The innovation of this paper is as follows.

1. We propose the idea of building multi-scene graphs and using ontologies to describe rich relation types and events.

2. We propose multi-scene graph merging, which combines multiple scene graphs to build a complete graph for joint analysis and inference.

3. We propose the entity-relation-event game rule controller to impose constraints on entity behavior.

\section{Related Work}
The Metaverse system \cite{sparkes2021metaverse} is mainly used in three scenarios. In terms of content distribution platforms, applications mainly cover the fields of games, social networks, sports, etc. In terms of display platforms and devices, it mainly includes smart wearable devices, VR, and AR. In terms of software and hardware technical support, it mainly includes blockchain, 5G networks, artificial intelligence (AI), cloud computing, chip design, etc.

There are some well-known metaverse systems in the world. There are different services in Facebook's Metaverse platform \cite{sparkes2021metaverse,wilson2012review}, covering social networks, games, work scene, collaboration, productivity, etc. Microsoft's Metaverse system \cite{kim2021advertising} includes digital twin-based games, augmented reality glasses, a mixed reality collaboration platform, cloud computing services, etc. Roblox is committed to providing users with the tools and technologies to create virtual worlds \cite{lee2021metaverse}. The content covers educational innovation, games, etc. Epic Games \cite{kaur2021metaverse} is working on building the metaverse into game systems. They build a high-fidelity digital human platform based on Unreal Engine in the cloud. Different from the above platforms, our metaverse framework aims to build a multi-scene metaverse and add entity-relation-event game rules to constrain the behavior of entities. 

\section{Approach}
This section explains the mechanism of the MetaOnce. We first define the notation of the framework. MetaOnce framework $M = (G, O, T)$ is composed of a graph $G = (V,E)$, an ontology $O = \{c_p,r^o_s,c_q\}$ and some events $T=\{e_t\}$. Graph is composed of vertices (nodes) $V=\{v_i\}$ and edges $E=\{(v_i,r^o_s,v_j)\}$. $c_p$ is a concept. $r^o_s$ is a relation defined in the ontology. $e_t$ is an event. $v_i$ is an entity. Our goal is to construct a metaverse framework to describe rich relations, events and game rules of entity-relation-event.

Our main effort lies in designing a portable metaverse framework that can construct a multi-scene metaverse. The main improvements consist of 3 aspects, the ontology, the multi-scene graph, and the game rules of entity-relation-event. In the following, We first describe the elements of the MetaOnce system. Then, we describe the concepts, relations, events. Then we describe the multi-scene graph merging. Finally, we describe the game rules for Entities, Relations, and Events.

\subsection{Elements of MetaOnce}
\begin{figure}[htbp!]
\centering
\includegraphics[width=2.5in]{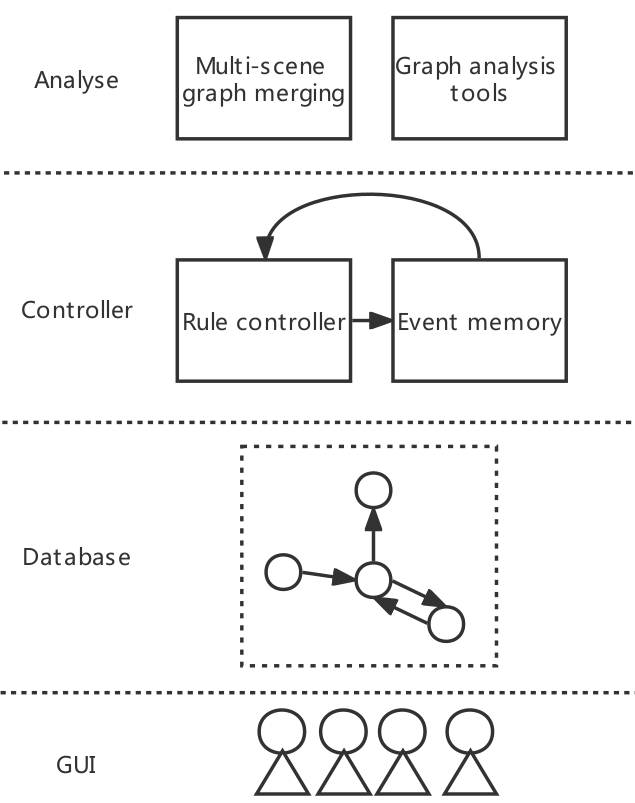}
\caption{System Architecture Overview\label{arch}}
\end{figure}
Based on the MetaOnce framework, we build an online MetaOnce system for users to interact with. The system architecture is shown in Figure \ref{arch}. This system is divided into 4 layers. At the GUI layer, users can access the system through a GUI (browser). At the database layer, we store the relations in the metaverse in the form of a graph. At the Controller layer, we design the rule controller and event memory modules, which will be introduced in detail in subsection Game Rules for Entities, Relations, and Events. In the Analyse layer, we design multi-scene graph merging and graph analysis tools. The content of multi-scene graph merging will be described in detail in subsection Multi-scene Relation Graph Merging.

Vertices represent entities. Edges represent relations between two entities. Relations can be represented as a graph. Our MetaOnce framework supports atomic operations on graphs and is highly versatile. 
Functions for analyzing graph data are defined in the graph search tool. These functions can be used to analyze the structure of the graph, such as finding core vertices (whose neighbors are densely connected) and key vertices (articulation point/cut vertices). The graph search function is used to analyze indirect connections between vertices in the graph and recommend vertices. Specifically, graph analysis tools include graph search algorithms, such as graph traversal, single-source shortest path \cite{sniedovich2006dijkstra}, the shortest path between two vertices, all paths between two vertices, path evaluation algorithm, key vertices discovery \cite{west2001introduction}, etc. 


\subsection{Ontology of MetaOnce\label{ontolgoy}}
To leverage common concepts, relations, and events, we propose to introduce open ontologies. As an illustrative example, our system references the schema.org \cite{guha2016schema} ontology to describe concepts, relations, and events. The schema.org vocabulary is created by Google, Microsoft, Yahoo, and Yandex and developed by an open community process. Shared glossaries make it easier for webmasters and developers to decide on a schema and get the maximum benefit from their efforts. Its main goal is to standardize HTML markup for webmasters to create rich results on topics of interest. It is part of the Semantic Web project, which aims to make document markup code more readable and meaningful to both humans and machines.

Users can expand new concepts at will when building their metaverse systems, such as adding more personalized classes and more fine-grained relations. For example, as shown in the section Experiment, we add some fine-grained concepts, such as ``Beauty'' and ``Handsome guy'' as ``subClassesOf'' the concept /Thing/Person. For the relation expansion, we add ``Make'' (an entity, make, a product), ``Leader'' (an entity, lead, an organization), etc. We use the subclasses of /Thing/Action as relations, such as BefriendAction, FollowAction, CommunicateAction, BelongsTo, MarryAction, JoinAction, LeaveAction, etc. For the event expansion, we add ``Divorce'', ``Sell'' events.


\subsection{Multi-scene Relation Graph Merging\label{mergesec}}
%
The relations that users establish in a particular scene no longer fit after moving to another scene. For example, the relations established in the classroom scene includes classmates, class representatives, monitor, and teacher-student. When switching to the game scene, the relations becomes teammate, opponent, assist, kill, heal, etc. A separately constructed scene graph makes it impossible to jointly analyze relations in multiple scenes. For example, A and B hold a relation (A, kill, B) in the game scene. In the classroom scene, B is A's teacher. Through the joint analysis of the two scenes, we can infer that A and B hold both a teacher-student relation and a friend relation.

Our framework can be used to build multilingual multi-scene graphs. When there are many scene graphs, we hope to merge these separate scene graphs into a single complete scene graph. Multi-scene graph merging aims to merge separately established scene graphs into a whole for analyzing more comprehensive entity-relation-event data through graph search methods. The difficulty in merging multi-scene relation graphs lies in aligning entities in two scene graphs and obtaining a union of relations. For the problem of aligning entities, our framework supports aligning entities based on entity IDs in multiple scenes. For the problem of relation union, since there can be multiple relations between two entities, the method we take is to traverse the relations of different scenes.

\subsection{Game Rules for Entities, Relations, and Events\label{rulesec}}
Establishing relations in the real world follows certain constraints. For example, if A buys B, no one else can buy the same B repeatedly unless A agrees to sell B. We design rules that impose constraints on the entity-relation-event game in the system. The challenges of designing this rule controller lie in that when the relations in the system change (establish, cancel, update), specific rules need to be triggered, and at the same time, the events that occur are persisted in the system. This ensures that every time the system restarts, it remembers previous events. For each request, the rule controller and event memory module will determine whether a relation can be established.

Implementing the metaverse's entity-relation-event game rules has two challenges. One is to intercept all requests in the system and determine whether each request follows the system rules. The other is to keep in mind the events that occur in the system to support the judgment of the rules. We propose a rule controller and an event memory module. The rule controller is designed to verify whether a user's request complies with the rules, and to impose or remove constraints from the rules to ensure that the system operates in compliance. The event memory module is a memory area in a metaverse system, which is specially used to record and retrieve events that occur in the Metaverse system, to ensure the compliant operation of the rule controller. 

The rule constraints we add are shown in Table \ref{constraint}, including exclusive constraints, symmetric constraints, asymmetric co-occurrence constraints, mutual termination constraints, and irreversible constraints.

\begin{table*}[!htbp]
\centering
\tiny
\resizebox{\linewidth}{!}{
    \begin{tabularx}{\linewidth}{c|p{3cm}|X|X}
    \toprule
    Type & \multicolumn{1}{c|}{Rule} & \multicolumn{1}{c|}{Description} & \multicolumn{1}{c}{Instances}  \\ 
    \midrule \hline
    Exclusive constraints &  $(A,r,B)\xrightarrow[]{}\neg(C,r,B)$ & If A has relation $r$ to B, then C cannot have relation $r$ to B & 1.If A and B are spouses, then C cannot be spouses with B
    \newline 2.If A buys B, then C cannot buy B \\ \hline 
    Symmetry constraints & $(A,r,B)\xrightarrow[]{}(B,r,A)$ &If A has a relation $r$ to B, then B also has a relation $r$ to A & 1.If A and B are spouses, then B and A are also spouses  \\ \hline
    Asymmetric co-occurrence constraints & $(A,r,B)\xrightarrow[]{}(B,r',A)$ where $r \neq r'$ &If A has relation $r$ to B, then B has relation $r'$ to A, where $r$ is not equal to $r'$ & 1.If A buys B, then B belongs to A  \\ \hline
    Mutual Termination constraints & $\neg(A,r,B)\xrightarrow[]{}\neg(B,r',A)$ & If A cancels relation $r$ with B, then B also cancels relation $r'$ & 1.If A sells B, then B no longer belongs to A \\ \hline
    Irreversible constraints & $\neg(A,r,B)\xrightarrow[]{}\neg(B,r',A)$, and A and B can no longer restore the relation $r$ and $r'$ & If A cancels relation $r$ with B, then B also cancels relation $r'$, and A and B can no longer restore the relation $r$ and $r'$ & 1.If A and B are divorced, then B and A are also divorced, and A and B cannot remarry each other  \newline 2.If A and B break up, then B and A break up, and A and B cannot get back together \\ 
    \bottomrule
    \end{tabularx}
}
\caption{Description and Examples of Some Rules\label{constraint}}
\end{table*}

\subsection{Experiments}
We first introduce the experimental setup and then conduct three experiments. First, we conduct scene graph construction experiments. Then, we conduct multi-scene graph merging experiments. Finally, we conduct experiments on the entity-relation-event game.
\subsection{Setup}
We evaluate system features using metaverse instances created by users based on their interests while the system is running online. The application is accessed through the chrome web browser. The system is deployed on an AMD Ryzen 5 1500X Quad-Core Processor @ Mem: 16G.

\subsection{Results of the Scene Graph\label{sec:scene}}
\begin{figure}[htbp!]
\centering
\subfloat[User Iron Man has logged in to the system \label{exp11}]{
\includegraphics[width=3in]{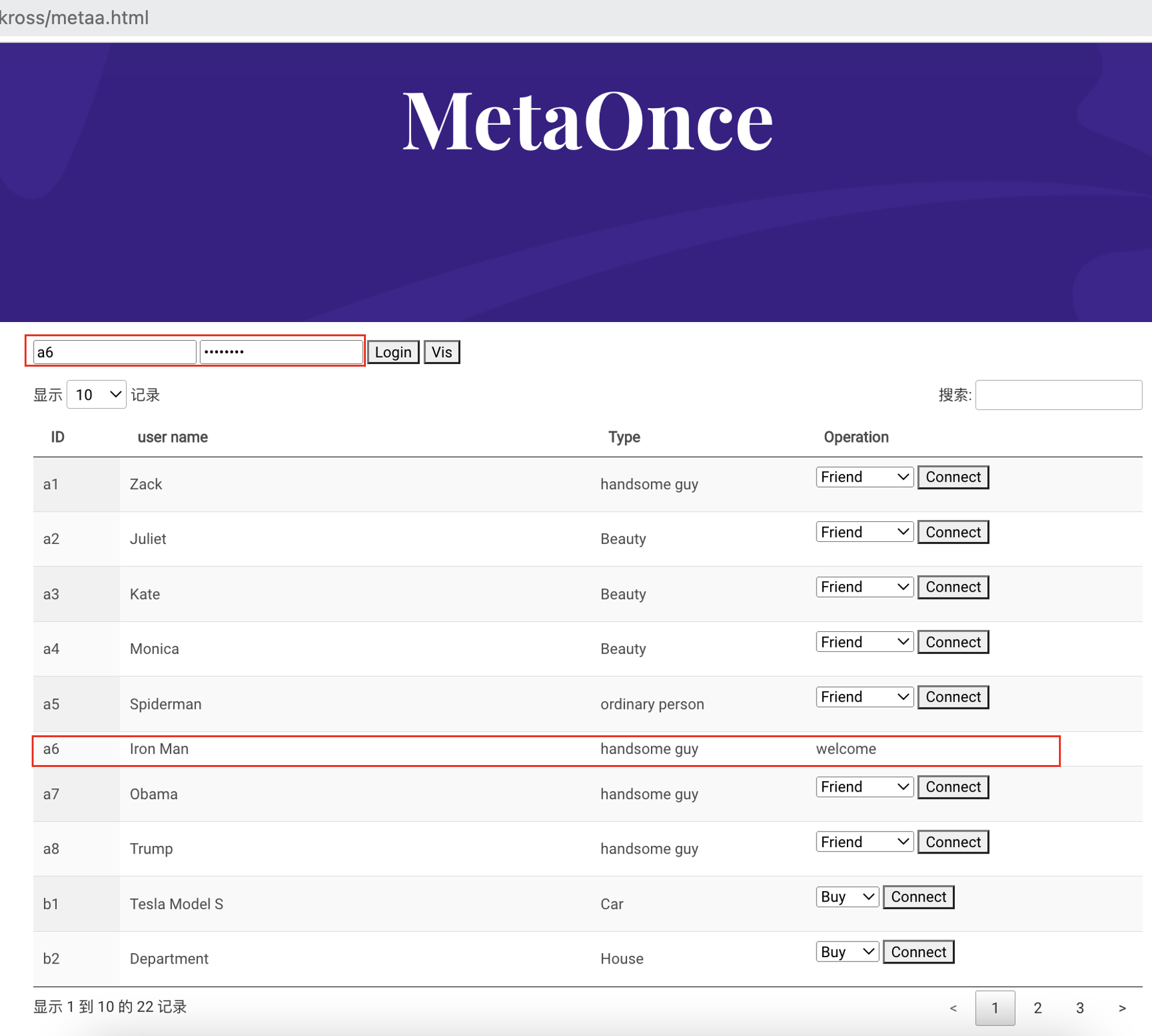}
}
\\
\subfloat[User Iron Man buys Weapon 1, adds Spiderman as a friend, leads the World peace department\label{exp12}]{
\includegraphics[width=3in]{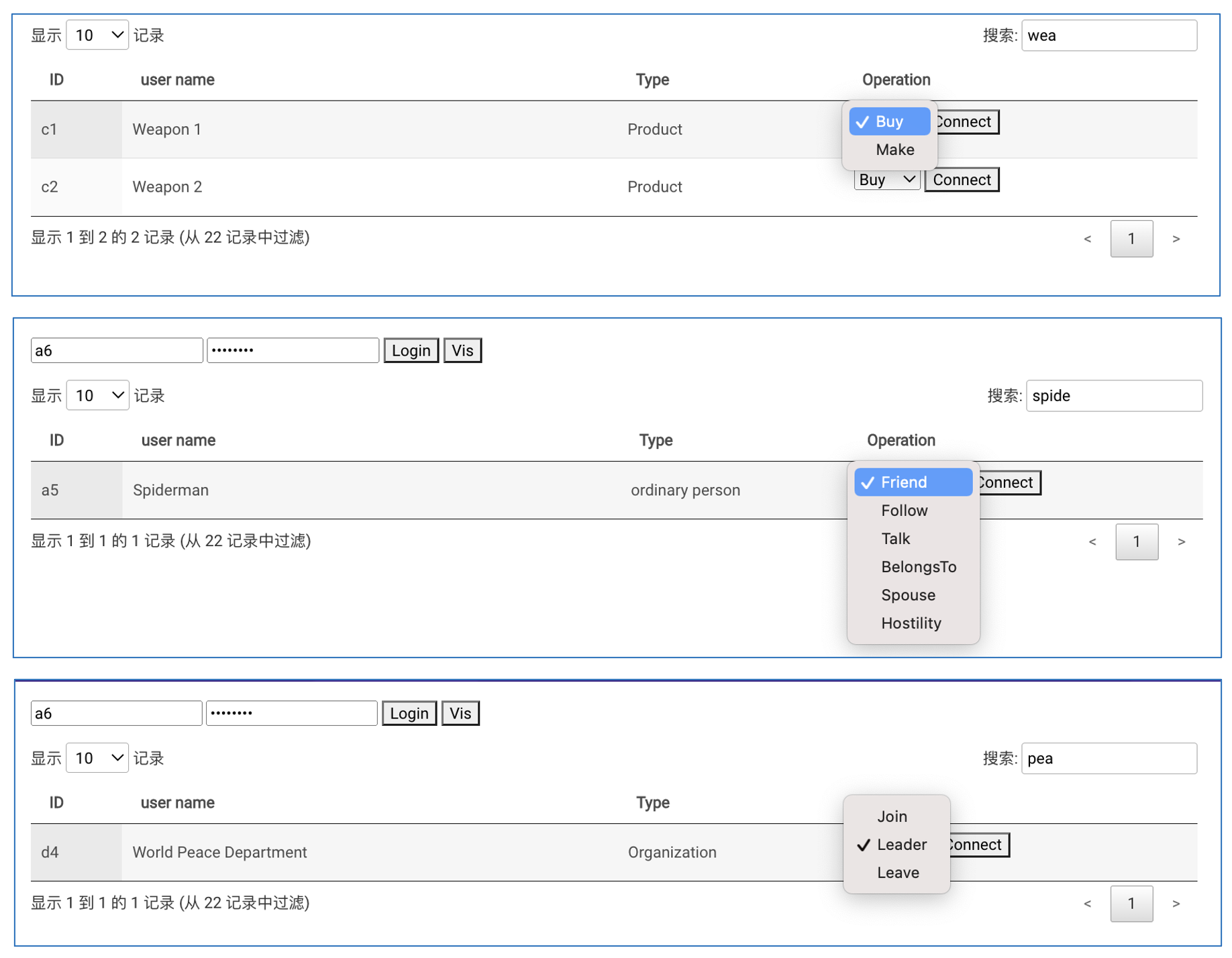}
}
\\
\subfloat[Graph visualization of the interstellar war scene\label{exp13}]{
\includegraphics[width=3in]{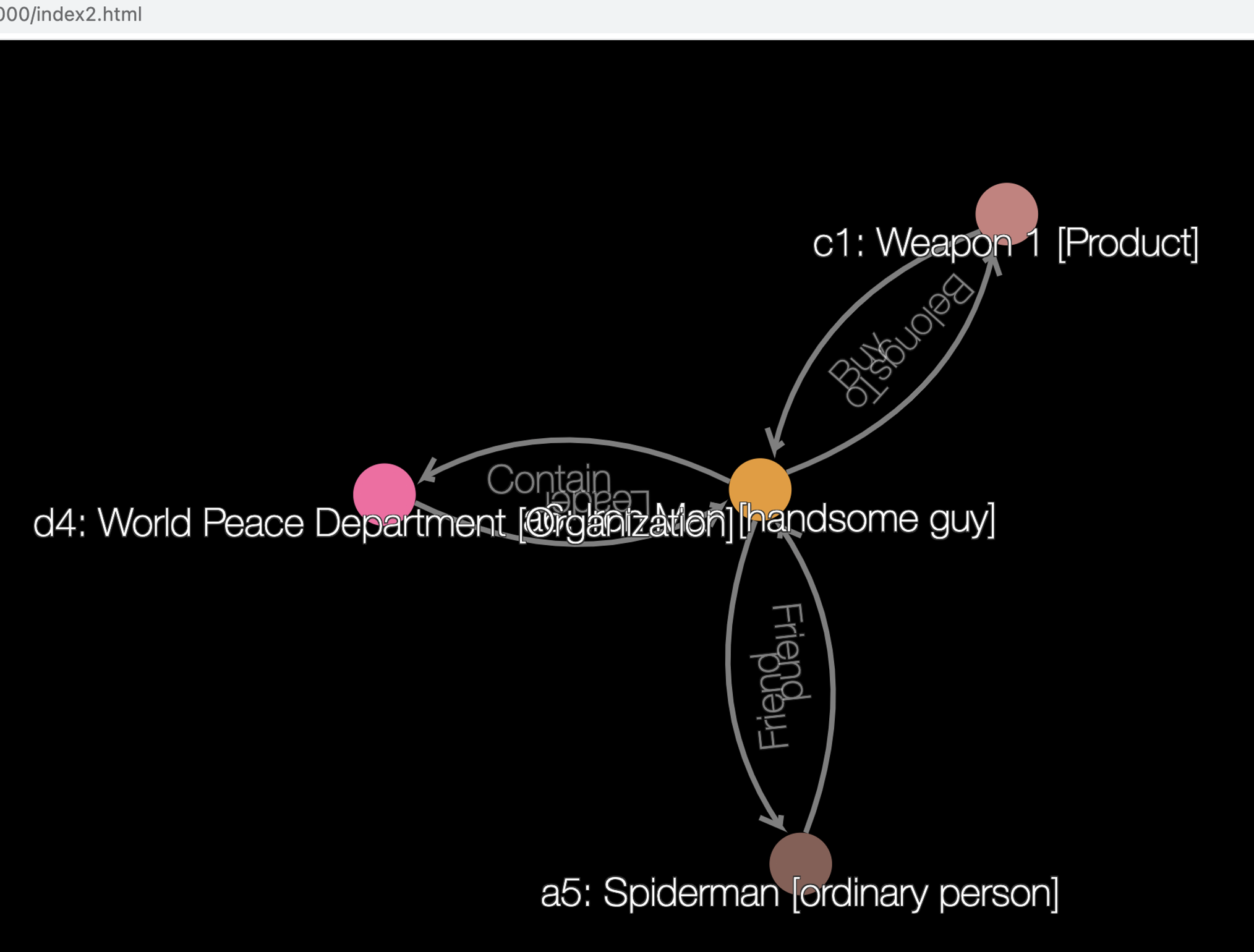}
}
\caption{Metaverse in an interstellar war scene \label{scene1}}
\end{figure}

This experiment evaluates building a scene graph on the MetaOnce system. This user-built scene graph includes rich relation types. The building process is shown in Figure \ref{scene1}. We set this scene as an interstellar war, and the red box in Figure \ref{exp11} represents the logged-in user. Users can establish or cancel relations with other entities in the metaverse. As shown in Figure \ref{exp12}, user Iron Man (a6) buys Weapon 1 (c1), adds Spiderman (a5) as a friend, leads the world peace department (d4). As shown in Figure \ref{exp13}, the relations between entities in this scene can be visualized. Different colored vertices in the graph represent different types of entities. Arrows represent directed relations between entities. 

\begin{figure}[htbp!]
\centering
\subfloat[User Spiderman joins the Classroom 1, then follows beauty Kate \label{exp21}]{
\includegraphics[width=3in]{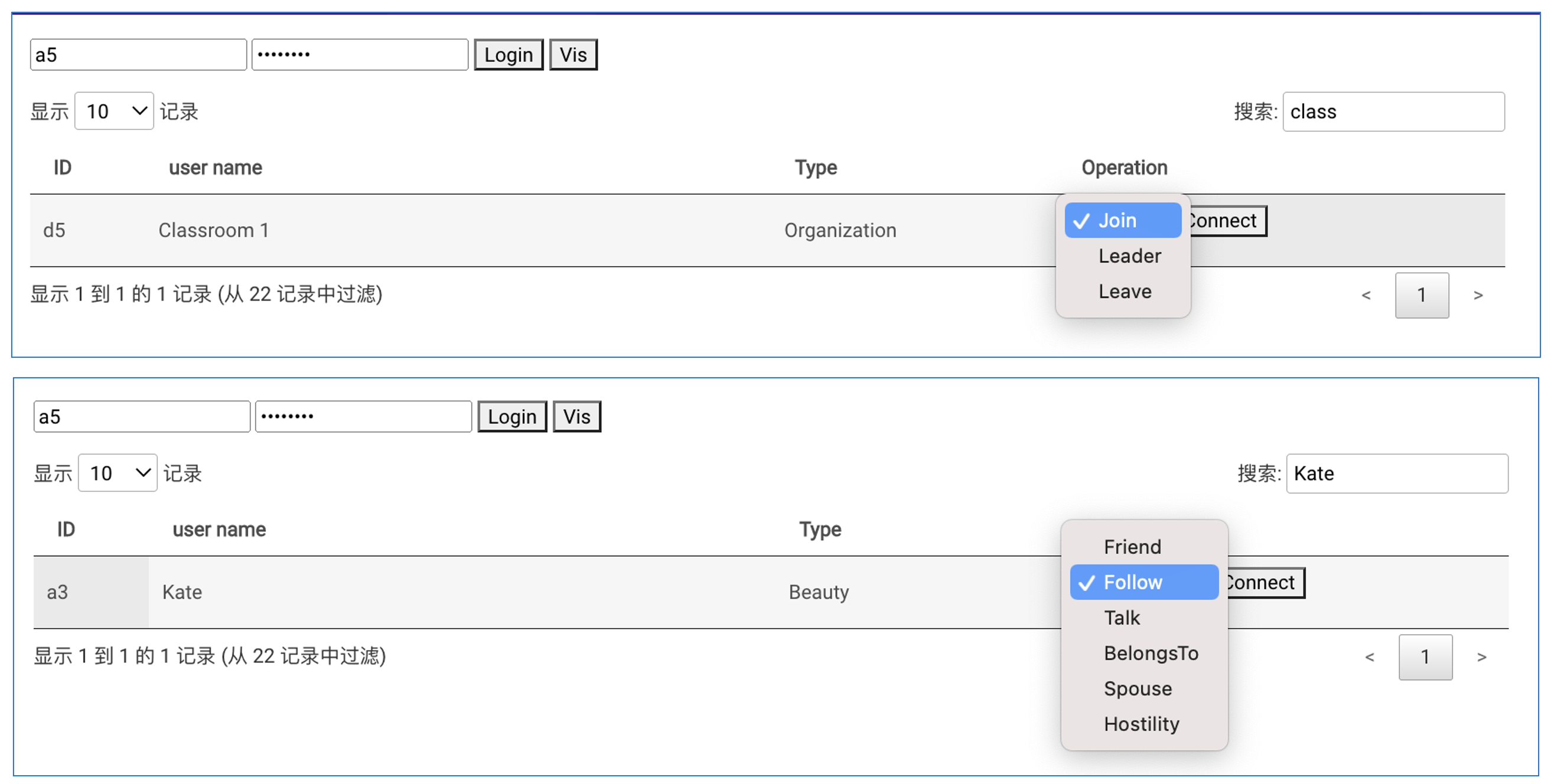}
}
\\
\subfloat[Graph visualization of the classroom scene \label{exp22}]{
\includegraphics[width=3in]{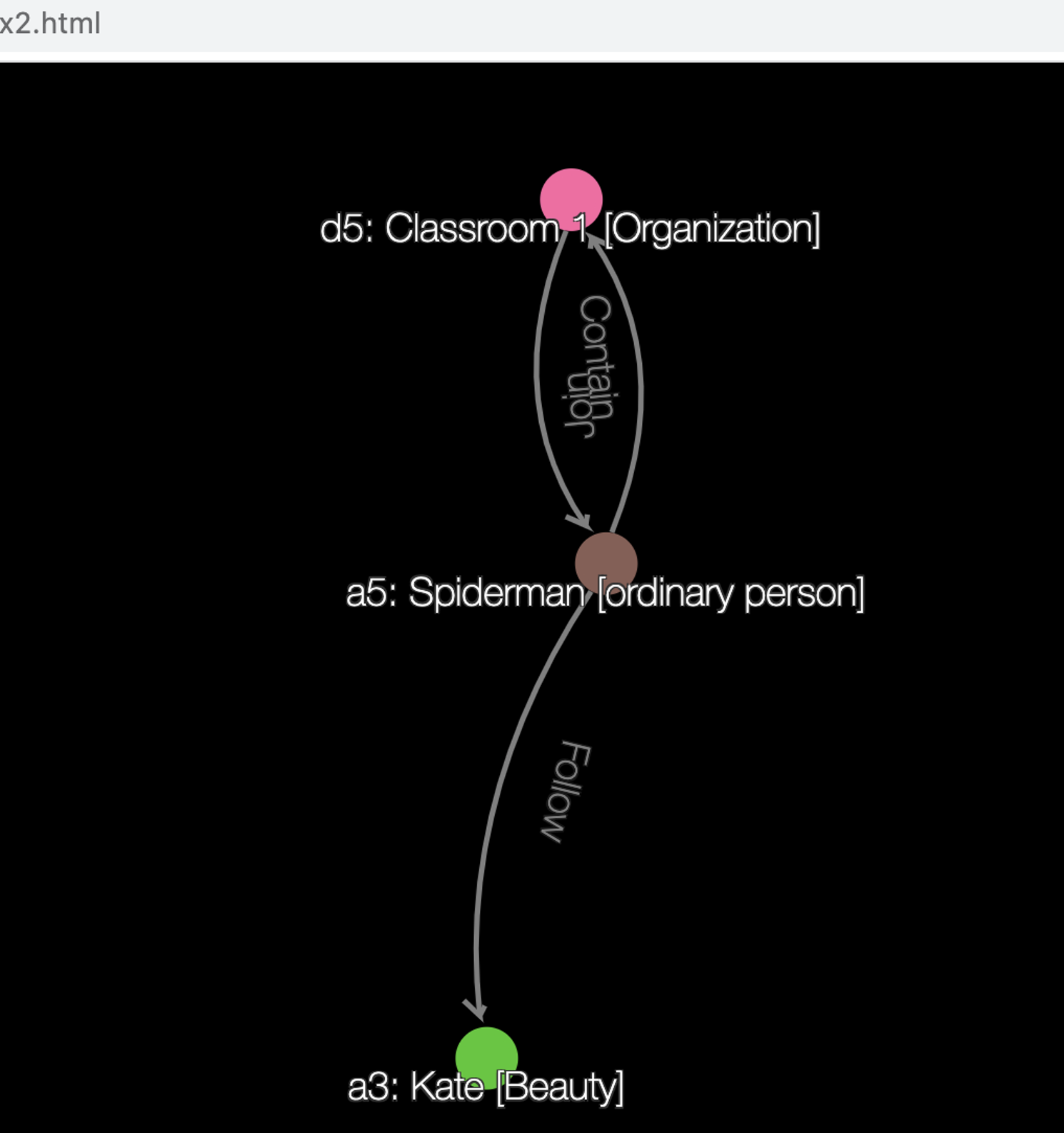}
}
\caption{Metaverse in a classroom scene \label{scene2}}
\end{figure}

The second scene is set to a classroom. User Spiderman (a5) logs into the system. As shown in Figure \ref{exp21}, User Spiderman (a5) joins Classroom 1 (d5), follows beauty Kate (a3). As shown in Figure \ref{exp13}, the relations between entities in this scene can be visualized. Different colored vertices in the graph represent different types of entities. Arrows represent directed relations between entities. 

The third scene is set to the Mars immigration. User Juliet (a2) logs into the system. As shown in Figure \ref{exp31}, user Juliet bought a car (b1) and a Mars farm (f1), and let the Mars farm (f1) buy a cow (e1). As shown in Figure \ref{exp32}, the relations between entities in this scene can be visualized. Different colored vertices in the graph represent different types of entities. Arrows represent directed relations between entities. 
\begin{figure}[htbp!]
\centering
\subfloat[User Juliet bought a car and a Mars farm, and let the Mars farm buy a cow\label{exp31}]{
\includegraphics[width=3in]{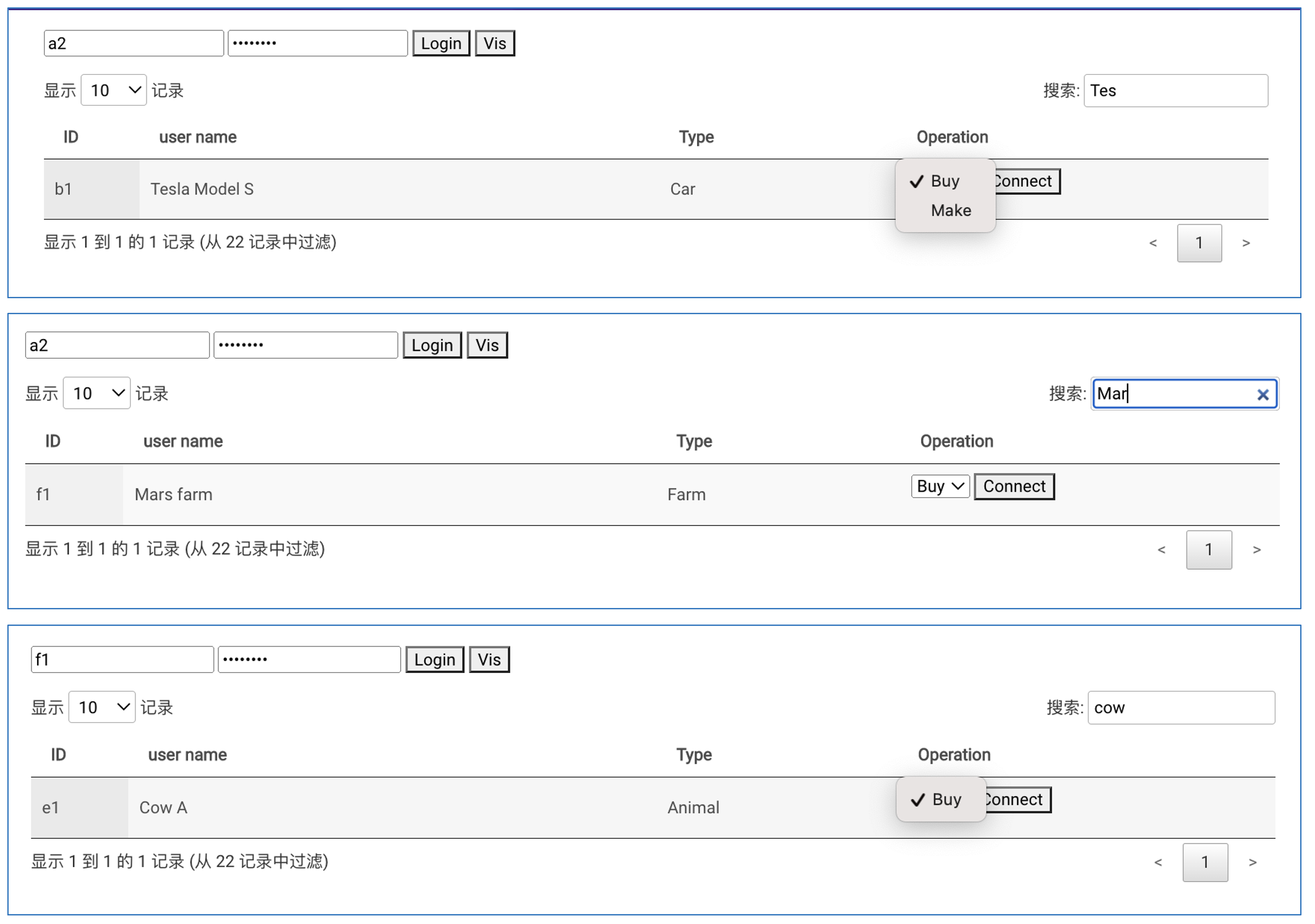}
}
\\
\subfloat[Graph visualization of the Mars immigration scene\label{exp32}]{
\label{2}
\includegraphics[width=3in]{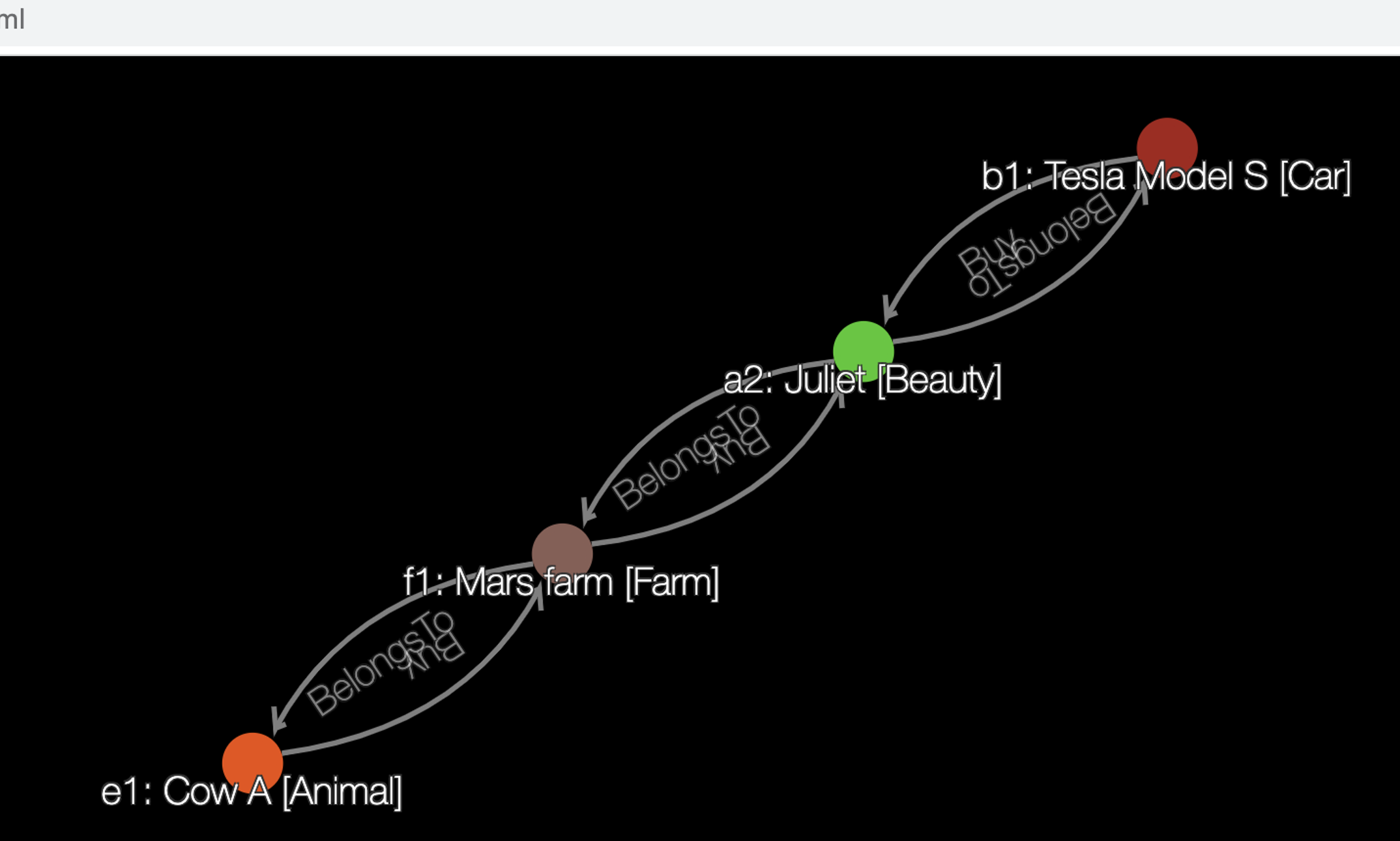}
}
\caption{Metaverse in the Mars immigration scene}
\end{figure}

\subsection{Results of Merging Multi-scene Graph}
This experiment evaluates the multi-scene graph merging. We merge the relations established in different scenes to get a graph describing the complete relations. Based on the experiments in subsection Results of the Scene Graph, we obtain 3 scene graphs. Then we merge the scene graphs, the result obtained by merging the first and second scene is shown in Figure \ref{exp31}. The result obtained by merging the 3 scenes is shown in \ref{exp32}. 
\begin{figure}[htbp!]
\centering
\subfloat[The result of merging the first two scenes\label{exp31}]{
\includegraphics[width=3in]{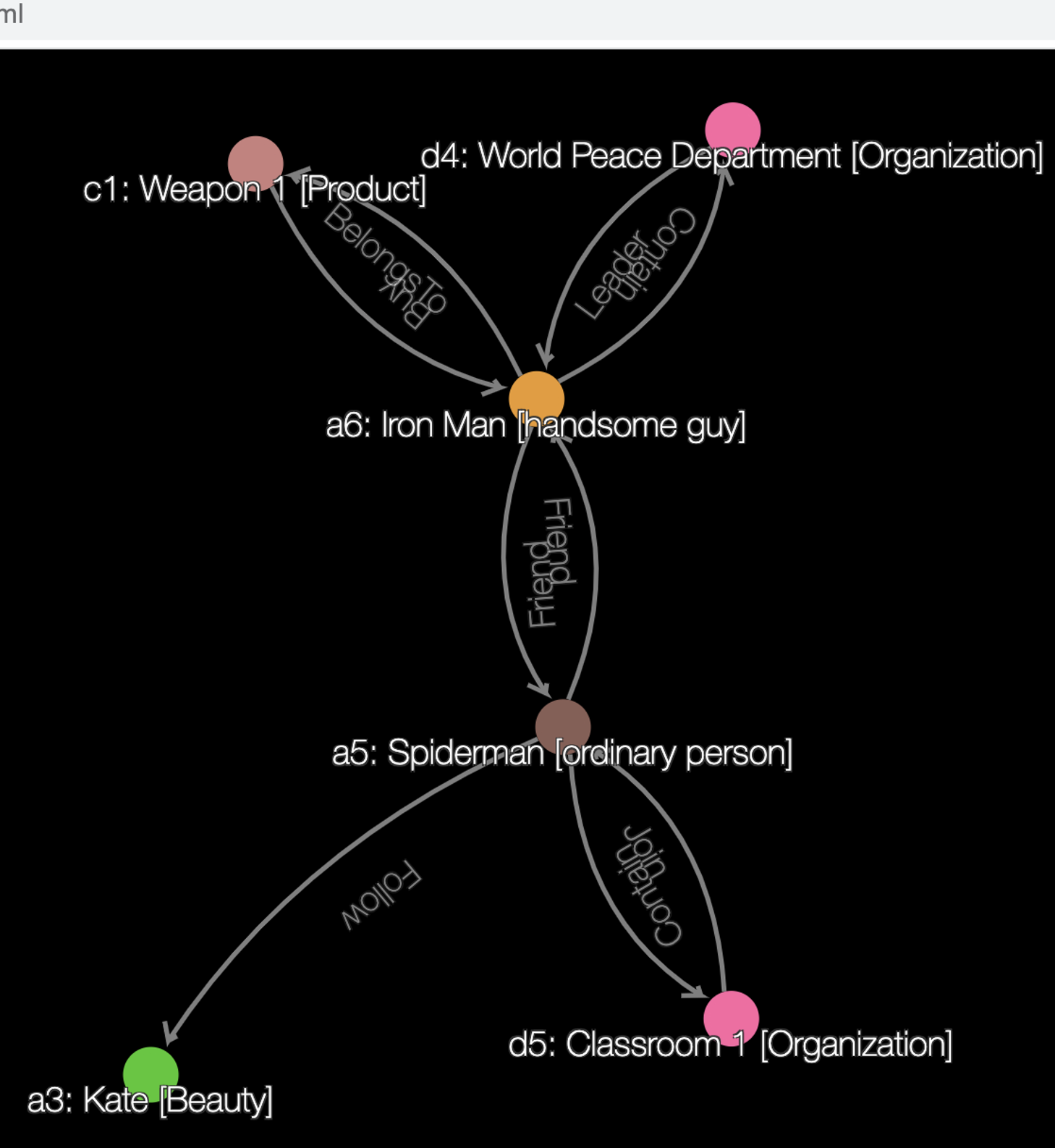}
}
\\
\subfloat[The result of merging all scenes\label{exp32}]{
\includegraphics[width=3in]{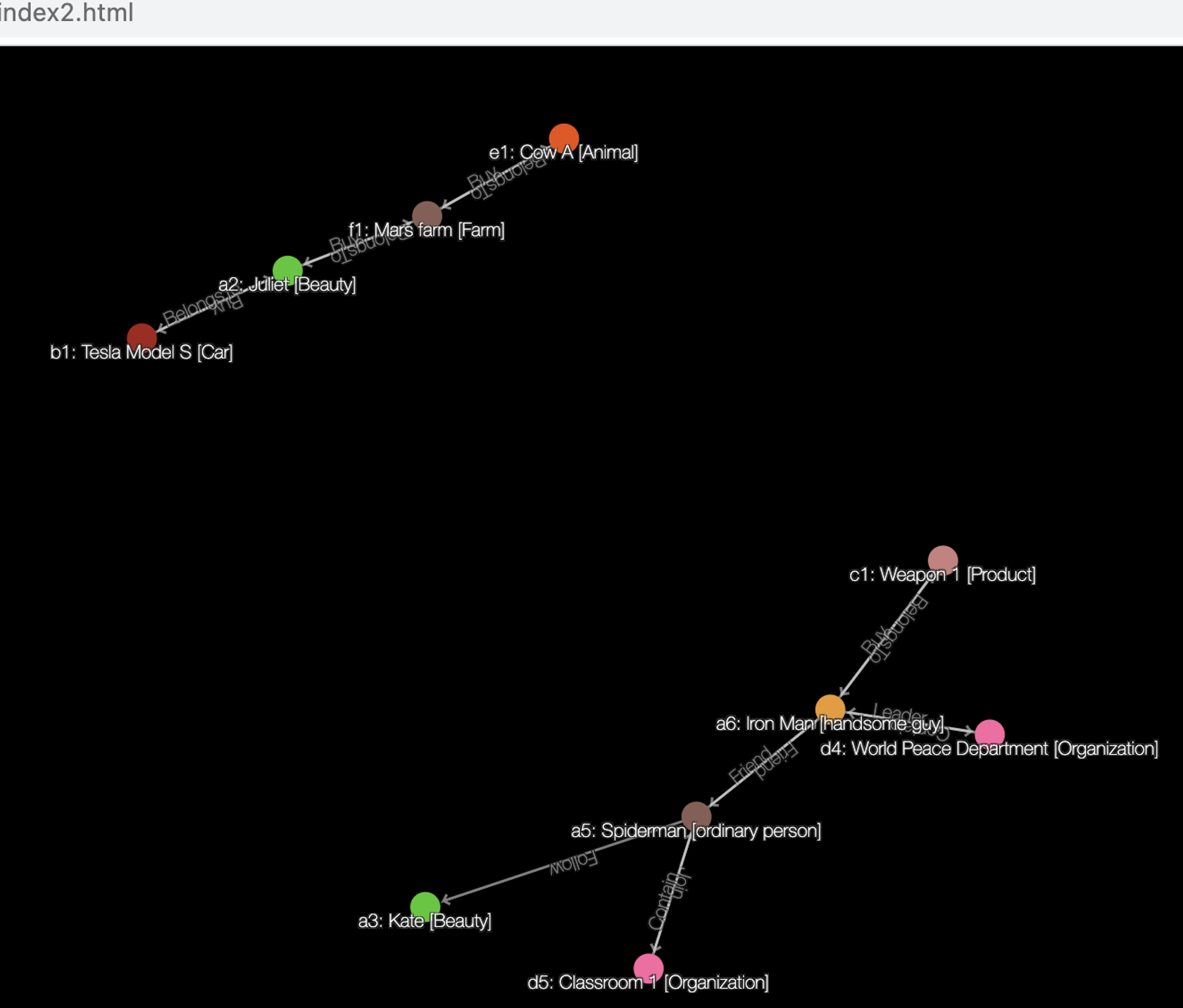}
}
\caption{Results of merging multiple scene graphs}
\end{figure}

\subsection{Results of the Entity-relation-event Game}
\begin{figure}[h]
\centering
\subfloat[Exclusive constraints\label{41}]{
\includegraphics[width=3in]{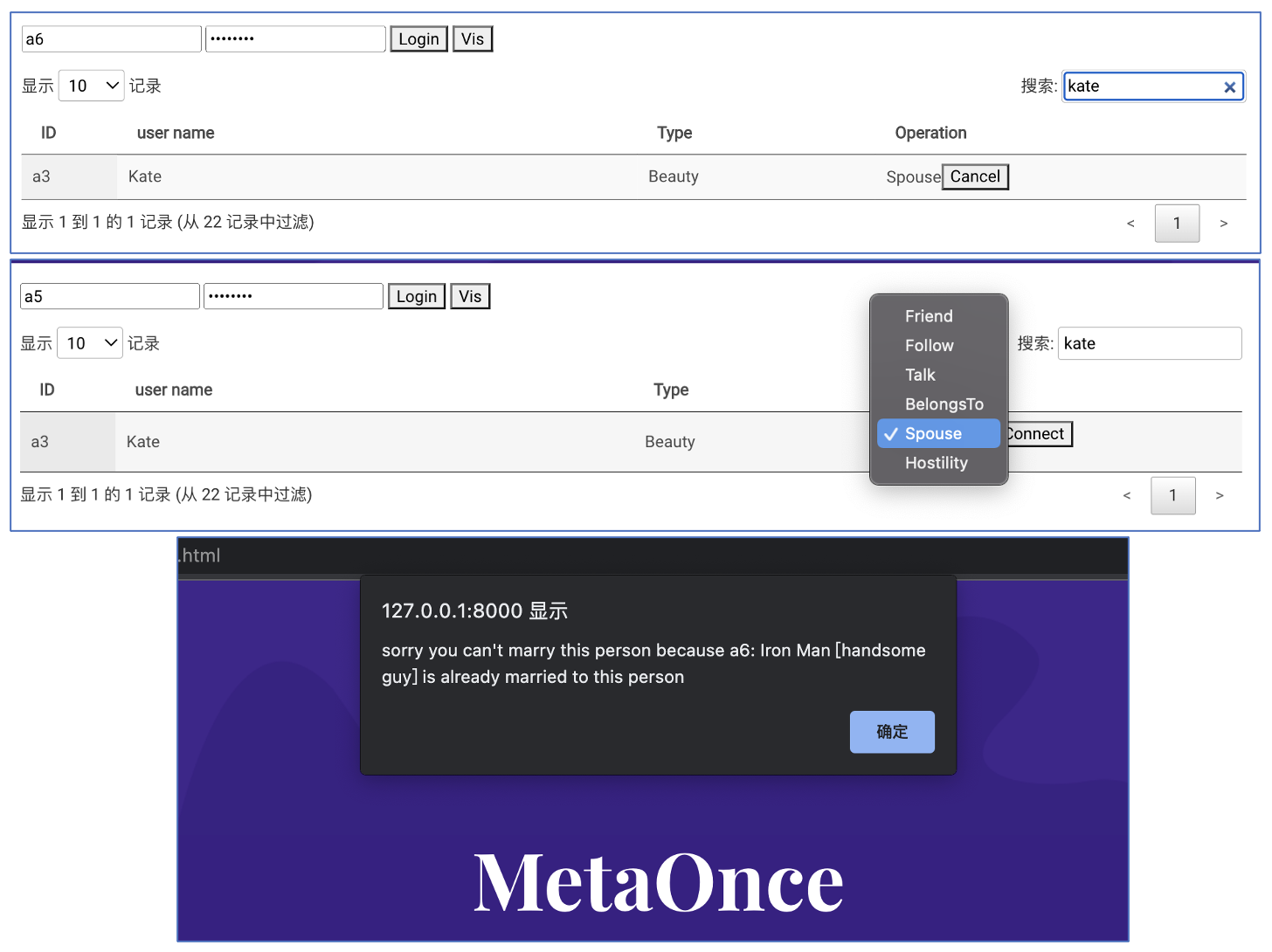}
}
\\
\subfloat[Exclusive constraints\label{42}]{
\includegraphics[width=3in]{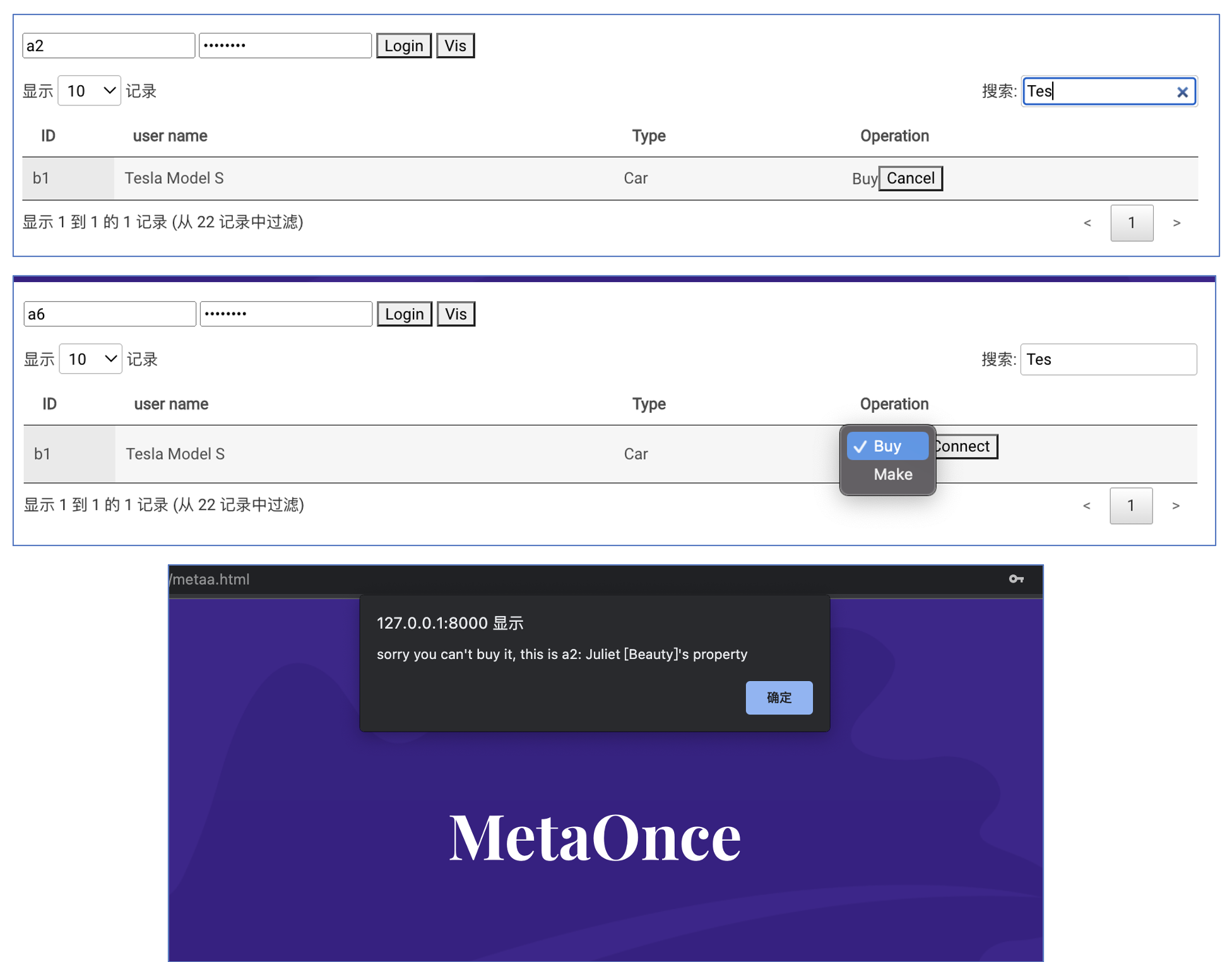}
}
\\
\subfloat[Irreversible constraints\label{43}]{
\includegraphics[width=3in]{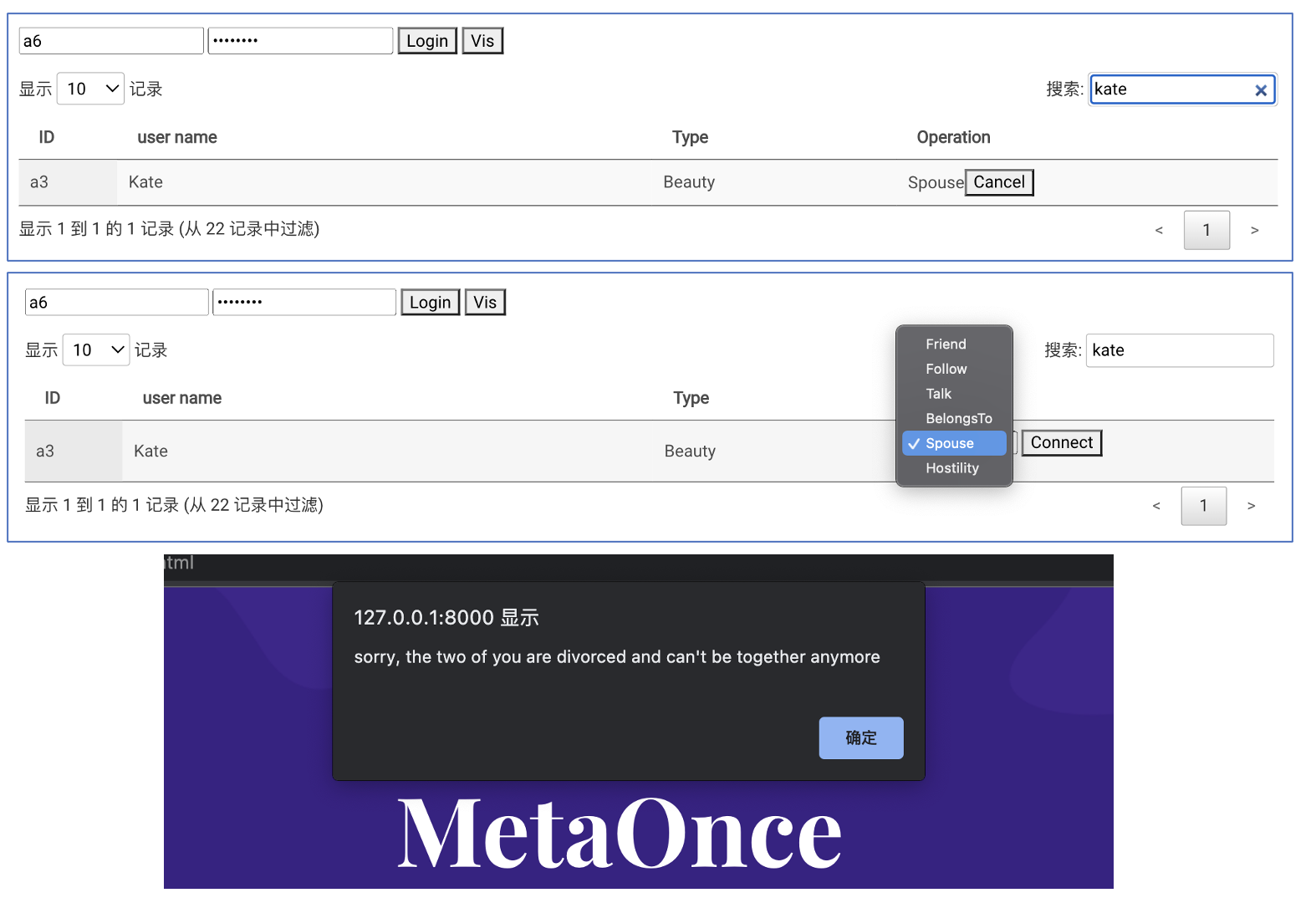}
}
\caption{Results of entity-relation-event game}
\end{figure}

This experiment evaluates the entity-relation-event game. Establishing or canceling a relation imposes constraints on the entity's behavior. Figure \ref{41} shows an example of exclusive constraints. Kate (a3) is married. When Spiderman (a5) logs into the system and wants to marry Kate (a3), the system will intercept the request and pop up a message ``Sorry, you can't marry this person because Iron Man is already married to this person''

Figure \ref{42} shows an example of exclusive constraints. Juliet (a2) bought a Tesla (b1). When Iron Man (a6) logs into the system and wants to buy a Tesla (b1), the system will intercept the request and pop up a message ``Sorry, you can't buy it, it's Juliet's property.''

Figure \ref{41} shows an example of irreversible constraints. Iron Man (a6) and Kate (a3) were previously married, but Iron Man (a6) logs into the system to divorce Kate (a3). When Iron Man chose to marry Kate again, the system will intercept the request and pop up a message ``Sorry, the two of you are divorced and can't be together anymore.''

\section{Conclusion and Future Work}
This paper proposes a new metaverse framework, MetaOnce, which aims to describe rich relations between entities in multiple scenes. We propose a method to merge multi-scene relations and a rule controller for the entity-relation-event game. We test these features by having users build metaverses with our system.

The innovation of this paper lies in constructing a new metaverse framework and proposing the idea of the multi-scene graph and the rule controller that the metaverse system follows. In the future, we hope to further enrich the large-scale graph data analysis capacity and enrich ontologies and events.

\end{CJK*}
\bibliographystyle{aaai}
\bibliography{reference}

\end{document}